\documentclass[sigconf,screen]{acmart}

\usepackage[utf8]{inputenc}
\usepackage[T1]{fontenc} 
\usepackage{microtype}
\usepackage{graphicx}
\graphicspath{{./images/}}
\usepackage{subcaption}
\usepackage{booktabs}
\usepackage{multicol}
\usepackage{multirow}
\usepackage{balance}

\usepackage{amsmath}
\DeclareMathOperator*{\argmax}{arg\,max}

\setcopyright{acmlicensed}
\acmPrice{15.00}
\acmDOI{10.1145/3460319.3464801}
\acmYear{2021}
\copyrightyear{2021}
\acmSubmissionID{issta21main-p58-p}
\acmISBN{978-1-4503-8459-9/21/07}
\acmConference[ISSTA '21]{Proceedings of the 30th ACM SIGSOFT International Symposium on Software Testing and Analysis}{July 11--17, 2021}{Virtual, Denmark}
\acmBooktitle{Proceedings of the 30th ACM SIGSOFT International Symposium on Software Testing and Analysis (ISSTA '21), July 11--17, 2021, Virtual, Denmark}

\def\comment#1{{\itshape\color{red}}}
\def\hadcom#1{{\itshape\color{green}}}

\begin{document}

\title{Exposing Previously Undetectable Faults\\ in Deep Neural Networks}
\renewcommand{\shorttitle}{Exposing Previously Undetectable Faults in Deep Neural Networks}

\author{Isaac Dunn}
\email{isaac.dunn@cs.ox.ac.uk}

\affiliation{%
  \institution{University of Oxford}
  \country{UK}
}

\author{Hadrien Pouget}
\affiliation{%
  \institution{University of Oxford}
  \country{UK}
}

\author{Daniel Kroening}

\affiliation{%
  \institution{Amazon, Inc.}
  \country{UK}
}

\authornote{The work reported in this paper was done prior to joining
	Amazon.}

\author{Tom Melham}

\affiliation{%
	\institution{University of Oxford}
	\country{UK}
}

\begin{abstract}

Existing methods for testing DNNs solve the oracle problem by
constraining the raw features (e.g.~image pixel values) to be
within a small distance of a dataset example for which the
desired DNN output is known.
But this limits the kinds of faults these approaches are able
to detect.
In this paper, we  introduce a novel DNN testing method that
is able to find faults in DNNs that other methods cannot.
The crux is that, by leveraging generative
machine learning, we can generate fresh test inputs that vary
in their high-level features (for images, these include
object shape, location, texture, and colour).
We demonstrate that our approach is capable of detecting
deliberately injected faults as well as new faults in
state-of-the-art DNNs, and that in both cases,
existing methods are unable to find these faults.
\end{abstract}



\maketitle

\section{Introduction}

As Deep Neural Networks (DNNs) begin to be deployed in safety critical and
mission-critical contexts, it becomes important to have confidence they are fit for
purpose. As with conventional software or hardware testing, evaluating a DNN
by checking its performance on an adequate suite of test inputs is a central approach
for revealing flaws present in the system.
Existing approaches generally focus on creating new test inputs by perturbing 
existing inputs. Methods for doing this with images include making small, undetectable changes 
to the pixels of an image~\cite{guo_dlfuzz_2018,lee_effective_2020}, using 
specific transformations such as translations 
or rotations~\cite{xie_deephunter_2019,gao_fuzz_2020,
Hosseini:SAE:2018,Engstrom:ETL:2019}, 
or adding noise and blur affects~\cite{Hendrycks:BNN:2019}. 
Beyond making changes directly to the pixels of an image,
more sophisticated methods leverage 
abstract representations of an input's 
features to make interesting changes~\cite{DBLP:conf/iclr/LiuTLNJ19,
DBLP:conf/nips/LiuBK17}.

Testing DNNs presents a slew of new challenges, relating 
to both the black-box nature of DNNs and their inherent 
design as approximations trained from finite data~\cite{DBLP:conf/safecomp/WillersSRA20}.
A particular challenge when testing DNNs is that there is
little sense of what properties the trained system is required
to satisfy, and so it is not always clear when a test result
indicates a violation of a required property. 

In fact, this specification problem is difficult even for 
the simplest imaginable specification: making the correct 
prediction on a given input. Not knowing the intended 
output for each input, the \textit{oracle problem}, is the 
reason DNNs are so valuable in the first place.
They are designed to generalise from labelled data to
make predictions on data for which \textit{we do not know the intended output}.
However, this makes testing them beyond the relatively small 
amount of labelled data we have difficult.
The simplest approach to testing DNNs does not solve the oracle problem
at all, instead reserving a set of manually labelled data to use solely
for testing. In this case, the only property that can be evaluated
is the accuracy on the original task, that is on data from exactly the
same source as the training data.
Existing methods that generate new dases for DNNs \emph{do} solve
the oracle problem. The solution is to restrict generated test inputs
to be sufficiently similar to a manually labelled example that we
can be confident that the desired system output is the same.
Most methods take one of two options.
First, and most commonly, test inputs are constrained to
have their raw features (e.g.~pixel values) differ no more than some fixed 
distance $\epsilon$ under an $\ell_p$ metric,
$\lVert x \rVert_p = (\Sigma_{i} \lvert x_i \rvert ^ p)^{\frac{1}{p}}$, typically with
$p = 2$ (Euclidean distance) or $p = \infty$.
Second, test inputs may differ in only specific hand-specified
ways from a manually labelled example. For instance,
by adding artificial fog or other corruptions~\cite{Hendrycks:BNN:2019},
by modifying the brightness or contrast of the input, or
by adding black squares~\cite{pei_deepxplore_2017}.

However, by introducing such constraints to circumvent the oracle problem, 
existing approaches significantly limit the
properties that they are able to evaluate.
For example, if generated test inputs are constrained to be
within an $\ell_\infty$ distance $\epsilon$ of known labelled
examples, then the tests are only able to expose faults that
fail to respect the invariant that changes of up to $\epsilon$
for each pixel should not change the system output.

In this paper, we introduce a testing method that is able 
to produce a much larger variety of tests than existing methods. 
While most existing methods can test for
invariance to $\ell_p$ $\epsilon$-bounded changes to pixel values,
or to certain hand-coded artificial feature changes,
our new approach is able to test for invariance to
changes to higher-level features such as position, colour, and texture of objects.
To do so, we leverage generative adversarial networks~\cite{Goodfellow:GAN:2014}, 
which have been trained
to encode the wide range of natural variation of features present 
in the data.

We demonstrate that our method is able to identify
particular faults that we intentionally create in DNNs,
and show that existing methods are unable to detect these.
We also apply our method to state-of-the-art image classification
DNNs, and again demonstrate that it is able to find faults
that cannot be found using existing approaches.

\section{Preliminaries}
\label{sec:background}
Generative adversarial networks (GANs)~\cite{Goodfellow:GAN:2014}
are a class
of generative machine learning models
involving the simultaneous training of two deep neural
networks: a {generator}~$g$ and
a {discriminator} $d$.
Specifically, given
a dataset $D \subseteq \mathbb{X}$ of samples drawn from
a probability distribution $p_D$,
the generator $g\colon \mathbb{Z} \to \mathbb{X}$
learns to transform
random noise $z$
drawn from a standard distribution
$p_z$
(typically Gaussian)
into an approximation of $p_D$.
The discriminator
network $d\colon \mathbb{X} \to \mathbb{R}$ learns to
predict whether
a given example $x$
is drawn from the data distribution $p_D$
or was generated by $g$.
The generator and the discriminator are \emph{adversarial}
because they train simultaneously, with each being rewarded
for out-performing the other.
That is, while the outputs of both are initially random,
the discriminator over time learns to identify features that
differ between the trained and generated data, which then
allows the generator to improve by adjusting that feature
of its generated data to match the training distribution.
Goodfellow's tutorial~\cite{Goodfellow:N2T:2017} can be
consulted for precise details.

A conditional GAN~\cite{Mirza:CGA:2014} is a variant that
learns to generate samples from a conditional
distribution
by simply passing the intended label $y$ for the
generated image to both the generator and
the discriminator, and training the generator
to maximise the log-likelihood
of the correct label in addition to
optimising its usual objective.
That is, a \emph{labelled} dataset
$D \subseteq \mathbb{X} \times \mathbb{Y}$ must be used
during training,
the discriminator
$d\colon \mathbb{X} \times \mathbb{Y} \to \mathbb{R}$
learns to discriminate between labelled dataset
and generated examples,
and the generator $g\colon \mathbb{Z} \times \mathbb{Y} \to \mathbb{X}$
learns to generate images with the specified label $y \in \mathbb{Y}$.

Generative adversarial networks have
one important property which makes them
especially suitable for test generation for image
classifiers: they
are able to learn to generate crisp high-quality
examples as though sampled
from a relatively complex training
distribution~\cite{Brock:LSG:2019}.
However, there are other approaches to
training generative DNNs---for instance
as a Variational Auto-Encoder (VAE)~\cite{Kingma:AVB:2014}---and
these would be perfectly suitable replacements
for a GAN-trained generator, if their performance
were satisfactory.

Generative networks have been found to display an 
interesting property: different layers, and even
different neurons, encode different kinds of features
of the image.
Earlier layers tend to encode higher-level information about 
objects in the image, whereas later layers
deal more with ``low-level materials, edges, and colours''~\cite[p.7]{Bau:GDV:2019}.
This makes sense:  the great promise of DNNs is their
ability to automatically construct hierarchies of feature 
representations.
In addition, by moving the input to the generator in a straight
line, features
such as zoom and object position and rotation
can vary in the image generated as its output~\cite{DBLP:journals/corr/abs-1907-07171}.
State-of-the-art GANs are particularly able to
smoothly and convincingly interpolate between different
images by so adjusting the input to the generator~\cite{Brock:LSG:2019}.
We will leverage these meaningful latent feature representations
when generating new test inputs.

\section{Our Method}
\label{sec:method}

The crux of our method is that by using a generative model
to \emph{generate}
fresh test data, rather than simply performing
small adjustments to existing test inputs,
it is possible to evaluate whether a DNN
behaves as required in response to variance
in features that
vary naturally in the training data.

\subsection{Problem Setup}

Suppose we have a set of possible system inputs $\mathbb{X}$, 
a set of discrete labels (system outputs) $\mathbb{Y}$, and an oracle 
$O\colon \mathbb{X} \to \mathbb{Y}$ that assigns to
each system input $x$ its `correct' output, $O(x)$.
When working with images, the input space is RGB pixel space
$\mathbb{X} = \mathbb{R}^{c \times w \times h}$,
where $c$ is the number of colour channels (typically $c = 3$), and
$w$ and $h$ are the width and height respectively of the image, in pixels.
For the task of object recognition, in which the system
is required to identify which of $k$ possible objects is depicted
in an image, the set of labels $\mathbb{Y} = \{1, 2, \ldots, k\}$
corresponds to the $k$ possible object classes.

Given a 
set of $N$ labelled datapoints $D = (x_i, O(x_i))_{i=1}^N \subset \mathbb{X} \times \mathbb{Y}$,
we can train a DNN image classifier
$f\colon \mathbb{X} \to \mathbb{R}^{|\mathbb{Y}|}$
that attempts to approximate $O$.
Given an input, $f$ outputs 
a real-valued confidence for each possible class $y \in \mathbb{Y}$.
Let $f_y(x)$ be the classifier's confidence of input 
$x$ being of class $y$, and $f_\mathit{pred}(x) = \argmax_y f_y(x)$, 
such that $f_\mathit{pred}$ is an approximation of $O$.
Typically, the final layer of a DNN is a softmax function, so that
for all output confidences $f_y(x)$, $0 \leq f_y(x) \leq 1$,
and
$\Sigma_{y=1}^k f_y(x) = 1$.

When testing a trained DNN, our task is to select test inputs
$x \in \mathbb{X}$. In particular, the goal is to identify
test cases that \textit{fail}, because these are indicative
of a fault in the DNN.
\begin{definition}
	A test case with test input $x \in \mathbb{X}$
	for DNN $f\colon \mathbb{X} \to \mathbb{R}^{|\mathbb{Y}|}$
	is said to \textit{fail} if
	$f_\mathit{pred}(x) \neq O(x)$.
\end{definition}

However, we quickly run into the
test oracle problem:
we do not have direct access to $O$ (if we did,
there would be no need to train an approximation $f$) and
it is too costly to seek human labelling for each test input.
So a practical test generation method needs to provide not
only test inputs~$x$, but additionally the desired system
output $O(x)$ so that failing test cases can be identified.

\subsection{Solving the Test Oracle Problem}
The standard approach to solving the test oracle problem
is to make use of the set $D$ of labelled data
that is available.
We can partition $D$ into a large training set
$D_\textit{train}$ and a small holdout test set
$D_\textit{test}$; by using only $D_\textit{train}$
during training, we can be confident that
the DNN has not overfit to any of the examples in $D_\textit{test}$,
so these examples can be used during testing.
For a test case 
$(x_\textit{test}, O(x_\textit{test})) \in D_\textit{test}$,
any new input $x_\textit{new} \in \mathbb{X}$ that we can
be confident shares the same desired output as $x_\textit{test}$
can therefore be used as a new test case, because
we simply assume that $O(x_\textit{new}) = O(x_\textit{test})$.

To identify such $x_\textit{new}$ that
share a label with a known test case,
most methods begin by choosing
a perturbation function
$t\colon \mathbb{X} \times \mathbb{P} \to \mathbb{X}$
with parameter space $\mathbb{P}$.
The intention is that, given a labelled test input 
$x_\textit{test}$, this function is able to generate
new test inputs as its parameter varies:
$x_\textit{new} = t(x_\textit{test}, p)$.
But to be confident that $x_\textit{new}$
is similar enough to $x_\textit{test}$ to have
the same true label, we must also introduce
a similarity constraint
over the parameter $p$.

\begin{definition}
	A \textit{similarity constraint} $d$ for a perturbation
	function $t$ is a function
	$d\colon \mathbb{P} \to \{\top, \bot \}$,
	such that for all $x \in \mathbb{X}$,
	if $d(p) = \top$ then $O(t(x,p)) = O(x_\textit{test})$.
\end{definition}

If we have a suitable perturbation function and similarity
constraint, then the problem of identifying test inputs
reduces to a search for suitable parameter values $p$:

\begin{proposition}
	If we have a labelled test case
	$(x_\textit{test}, O(x_\textit{test})) \in D_\textit{test}$,
	a perturbation function $t\colon \mathbb{X} \times \mathbb{P} \to \mathbb{X}$,
	a similarity constraint $d\colon \mathbb{P} \to \{\top, \bot \}$,
	and a parameter $p \in \mathbb{P}$,
	and if $d(p) = \top$ and
	$f_\mathit{pred}(t(x_\textit{test}, p)) \neq O(x_\textit{test})$,
	then $t(x_\textit{test}, p)$ is a failing test case.
\end{proposition}

\subsubsection{Test Generation using Pixel-Space Perturbations}
\label{sec:method:pixperts}
Most existing methods that generate tests
for DNNs use a pixel-space perturbation approach.
We have that $\mathbb{P} = \mathbb{X}$ and
$t(x, p) = x + p$. In effect,
$t$ simply changes each pixel value in the image independently.
The similarity constraint used typically constrains the
magnitude of $p$: if the pixel values do not change too
much, then the label should remain the same.
This magnitude is typically measured using the
$\ell_{\infty}$ or $\ell_2$ norm metric.
That is, $d(p) = \lVert p \rVert_2 \leq \epsilon$
or $d(p) = \lVert p \rVert_{\infty} \leq \epsilon$,
for a manually chosen constant $\epsilon$
small enough that the change is nearly imperceptible.

Given a labelled test case
$(x_\textit{test}, O(x_\textit{test})) \in D_\textit{test}$,
then, a pixel perturbation method must
find a suitable $p$. This is almost always done using an optimisation
over $p$, using
a loss function chosen to be
minimised when $f_\mathit{pred}(t(x_\textit{test}), p) \neq O(x_\textit{test})$
and $d(p) = \top$.
Since DNNs are differentiable, the derivative of the loss function with
respect to $p$ can be computed, and this gradient can be walked to minimise
the loss and thereby identify a failing test case.
If additional properties are desired of the test cases, this can be
reflected in the choice of loss function.

\subsection{Using GANs to Perturb Images}
In this paper, as in existing methods,
we use a perturbation-based approach
to solve the test oracle problem.
However, rather than directly perturbing the pixels of
a labelled test dataset image,
there are two important differences:
\begin{enumerate}
	\item Instead of beginning with a labelled test
	dataset image, we use a conditional generative
	network to \emph{generate} a fresh test seed
	for which we know the correct label.
	\item Instead of perturbing the individual
	pixel values of this seed, we make perturbations
	to meaningful features of the input by
	exploiting the generative network's learnt features.
\end{enumerate}

\subsubsection{Generating Test Seeds}

Suppose we have a pretrained conditional generator network
$g\colon \mathbb{Z} \times \mathbb{Y} \to \mathbb{X}$, which as described 
in Section~\ref{sec:background}, takes a normally sampled 
$z \in \mathbb{Z}$ and a class label $y \in \mathbb{Y}$ and returns an image 
$g(z, y) \in \mathbb{X} = \mathbb{R}^{c \times w \times h}$
such that $O(g(z, y)) = y$.
Now, rather than relying on the finite examples in a test dataset
as our source of seeds from which to create test cases,
we can generate fresh labelled test seeds on demand, by
sampling new generator inputs $z$.

While this may be valuable in itself, our main intention in
using generated images is that it allows much greater control
over the test inputs we create.

\subsubsection{Making Perturbations}
Because the generator $g$ is a deep neural network,
it can be described as a sequence of $n$ layers.
By writing the $i$th layer as a function
between layer outputs
$g_i\colon \mathbb{O}_{i-1} \to \mathbb{O}_i$, where $\mathbb{O}_0 = \mathbb{Z}$
for convenience
and $\mathbb{O}_n = \mathbb{R}^{c \times w \times h}$,
we can write $g$ as a function composition:
$g = g_n \circ g_{n-1} \circ ... \circ g_1$.

The thrust of our method is to introduce a perturbation
function that perturbs high-level features rather than
individual pixel values.
Since the neurons in a generative network encode meaningful
features~\cite{Bau:GDV:2019},
we perturb the activations of these neurons so as to adjust
the features of the generated image in a context-sensitive way.
That is, rather than using a perturbation parameter space
$\mathbb{P} = \mathbb{X}$,
our parameter space allows adjustments at the output
of multiple layers in the generator:
$\mathbb{P} = \mathbb{O}_0 \times \mathbb{O}_1 \times \ldots \times \mathbb{O}_n$.
Refer to our previous paper~\cite{dunn2020evaluating}
for a less concise exposition of this method.

Then we define our perturbation function
\begin{equation}
	t(g(z, y), p) = (g'_n \circ g'_{n-1} \circ ... \circ g'_1)(z, y, p),
\end{equation}
where $g'_i(o_{i-1}, p_{i-1}) = g_i(o_{i-1}) + p_{i-1}$.
That is, at each layer in the generator,
$t$ simply performs element-wise addition of the parameter tensor
with the layer output.
Our similarity constraint measures the total size of the changes
being made to the activations:
$d(p) = \lVert \bar{p} \rVert_2 < \epsilon$, where $\bar{p}$ is
the one-dimensional vector formed by flattening all the elements
of $p_0, p_1, \ldots, p_n$ into a list.

Our previous work~\cite{dunn2020evaluating} includes experiments
with human participants that verify that this similarity constraint
approach works well. These found that for around 80\% of generated
tests, the new test is judged to indeed have the same label as the
original test seed. This value can be increased at the cost of more
computation per test by simply increasing the effective $\epsilon$
constraint, if required, but for our purposes, 80\% is certainly
high enough to reveal faults in the models under test.
Note that these experiments also verify that the generator
is able to consistently generate images that humans would
correctly label as the desired class.

For our experiments in this paper, we only optimise over the first six layers
of the 18-layer generator, since the earlier layers
encode high-level, human-intelligible features~\cite{Bau:GDV:2019}.

\subsubsection{Finding a Suitable Perturbation}
As before, given a labelled test seed,
in this case $(g(z, y), y)$,
and our perturbation function~$t$ and similarity constraint~$c$,
finding a value $p \in \mathbb{P}$ suffices
to produce a failing test case for the DNN under test,~$f$.
We use a gradient-walking optimisation to find such satisfactory
values of $p$.
In particular, we use the loss function
$L(p) = \max_y f_\mathit{pred}(t(g(z, y), p))_y$,
which penalises confidence in the DNN's top output class.
It is also possible to include a penalty on $\lVert \bar{p} \rVert_2$,
but in practice we found this to be unnecessary:
by starting with every element of $p$ set to 0,
the gradient walk increases $\lVert \bar{p} \rVert_2$
slowly enough that it is acceptably small
when a suitable $p$ is found.
Note that the usual backpropagation algorithm is
sufficient to compute gradients of $L$ with respect to
$p$ since $f$, $t$ and each layer of $g$ are differentiable.

\newcommand{\addtableimg}[1]
{\raisebox{-.5\height}{\includegraphics[width=0.45\linewidth]{#1}}}
\newcommand{\addexample}[1]
{\addtableimg{examples/injected/#1_unpert_generated_x_grid_0.jpg}
	\!$\rightarrow$\!
	\addtableimg{examples/injected/#1_generated_x_grid_0.jpg}}
\newlength{\tablevgap}

\begin{table*}[]
	\caption{Examples of tests for DNNs with deliberately injected flaws.
		To the left of each arrow is the generated test seed which is correctly
		classified; to the right of each arrow is the generated
		test input that is incorrectly classified.
		Inspection of these test cases indicates that the DNNs
		are relying on the injected fault features
		(refer to Table~\ref{tab:fault-summary}).
	}
	\label{tab:injectedeg}
	\setlength{\tablevgap}{12mm}
	\begin{tabular}{lp{0.22\linewidth}p{0.22\linewidth}|p{0.22\linewidth}p{0.22\linewidth}}
		\toprule
		\# &\multicolumn{2}{c}{Tests that indicate the fault $y_0 \rightarrow y_1$ }  & \multicolumn{2}{c}{Tests that indicate the fault $y_1 \rightarrow y_0$}\\ 
		\midrule
		1 & \addexample{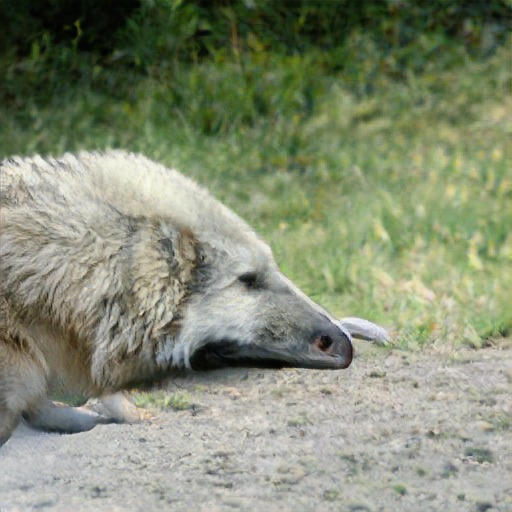} & \addexample{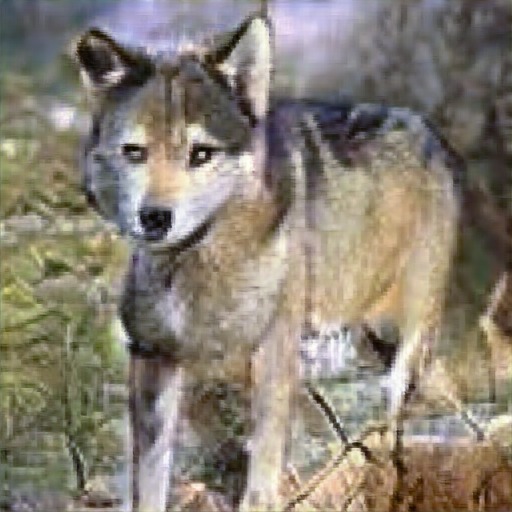} &
		\addexample{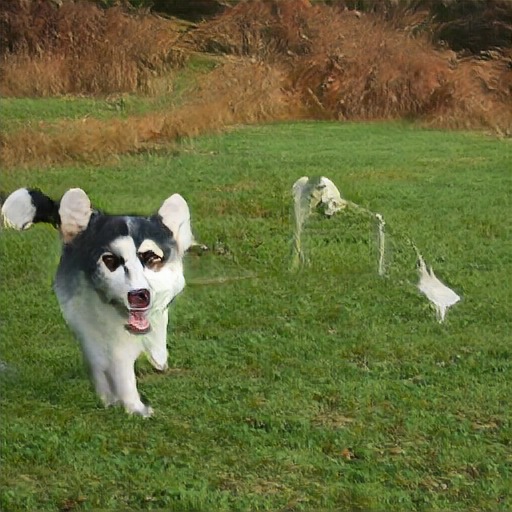}& \addexample{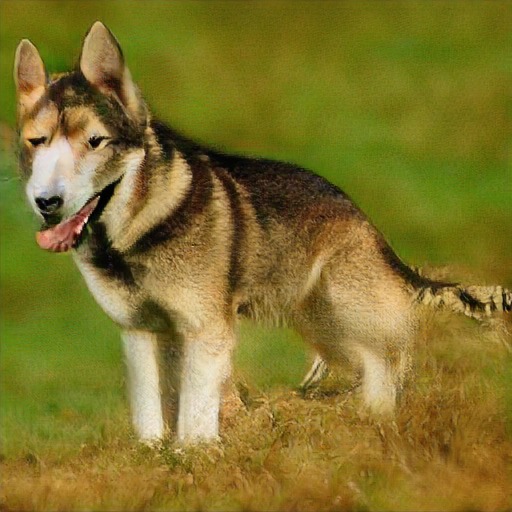} \\[\tablevgap]
		2 & \addexample{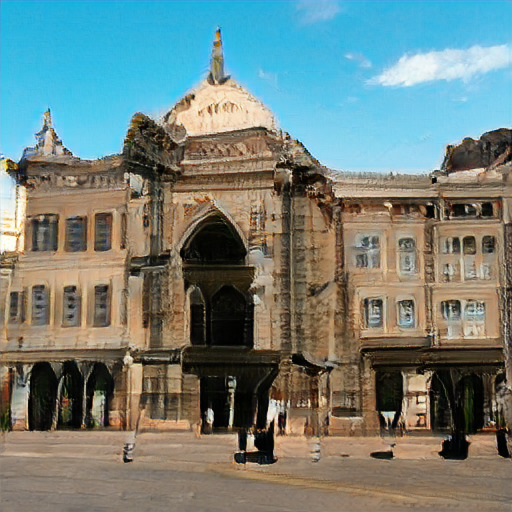} & \addexample{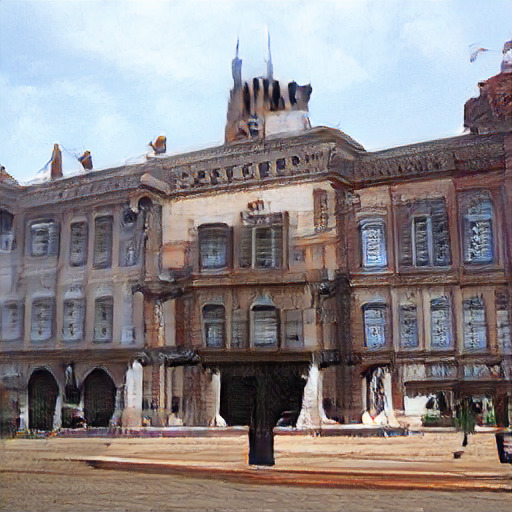} &
		\addexample{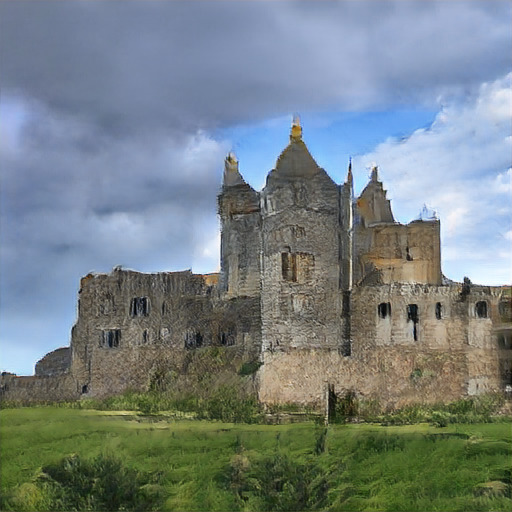} & \addexample{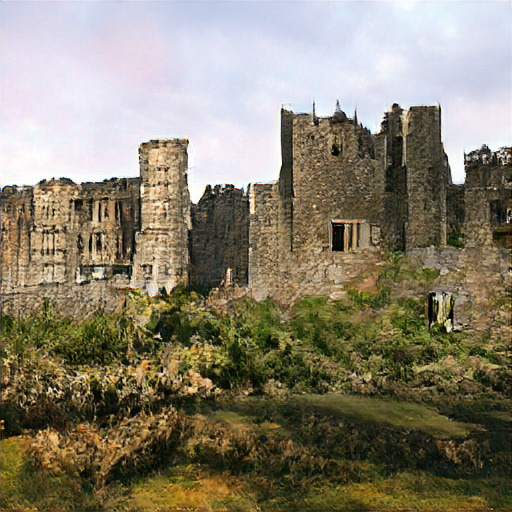}  \\[\tablevgap]
		3 & \addexample{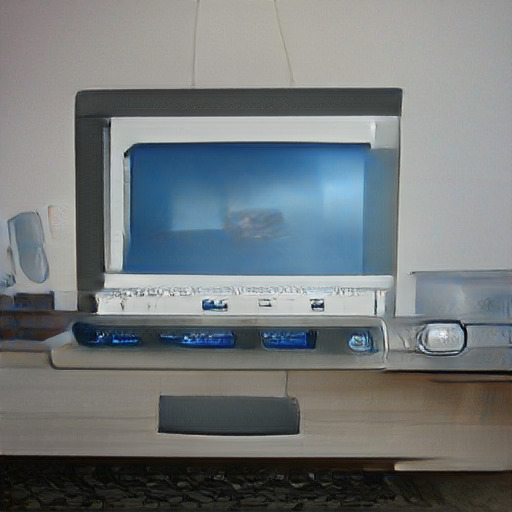} & \addexample{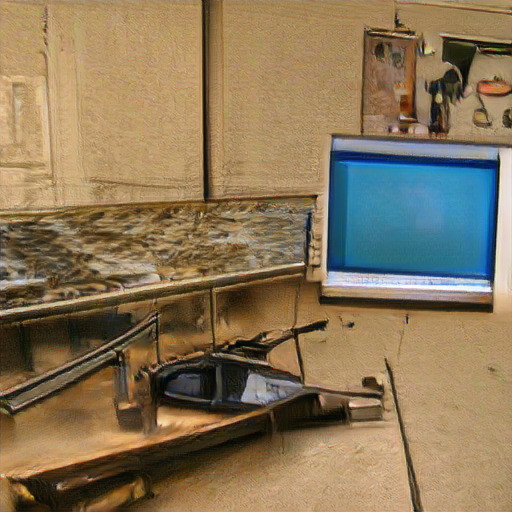} &
		\addexample{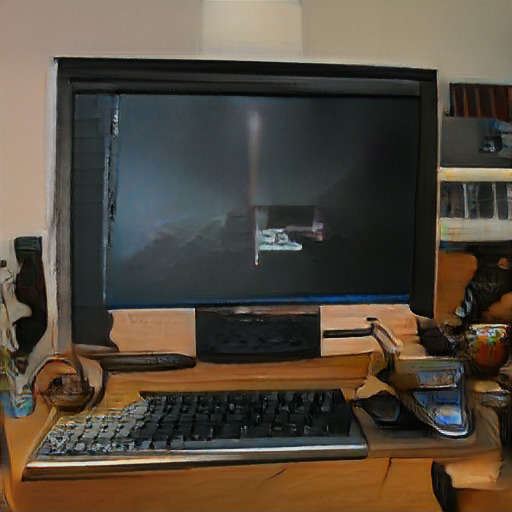} & \addexample{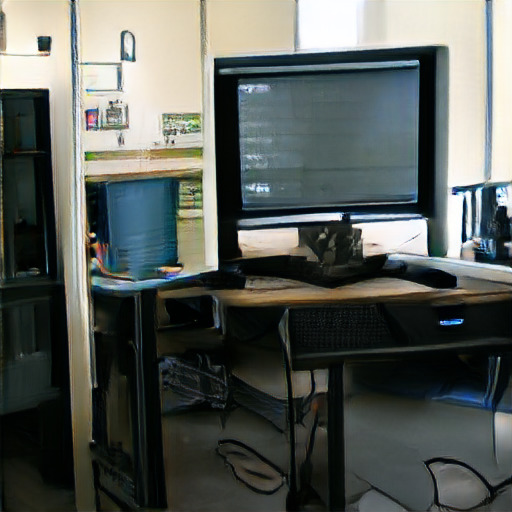} \\[\tablevgap]
		4 & \addexample{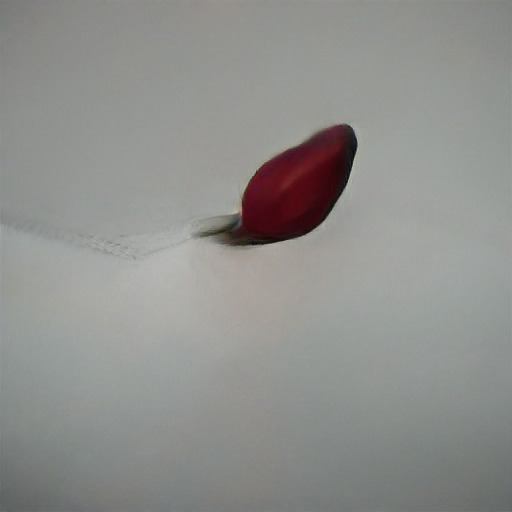} & \addexample{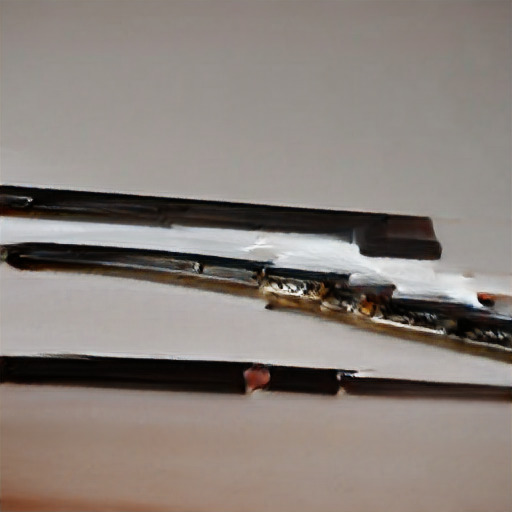} &
		\addexample{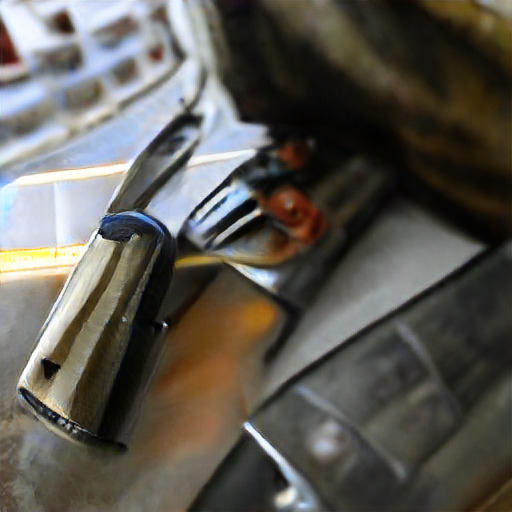} & \addexample{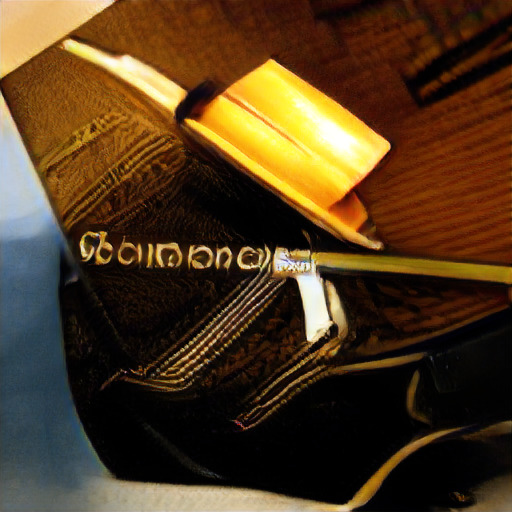} \\[\tablevgap]
		5 & \addexample{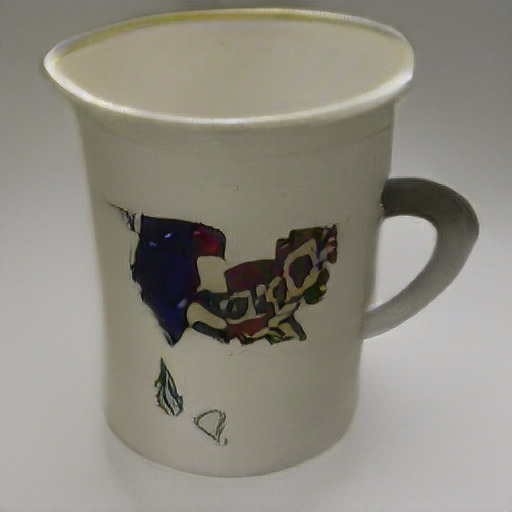} & \addexample{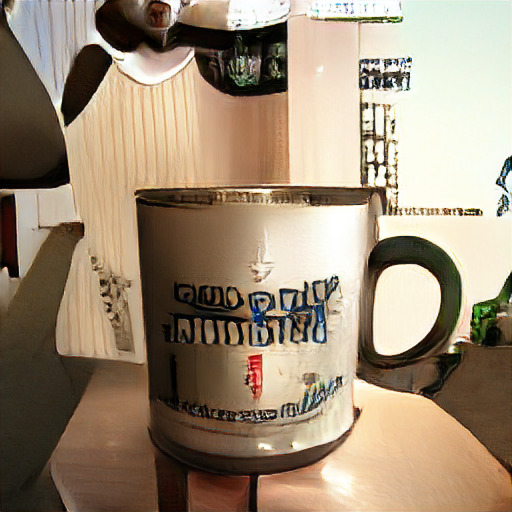} &
		\addexample{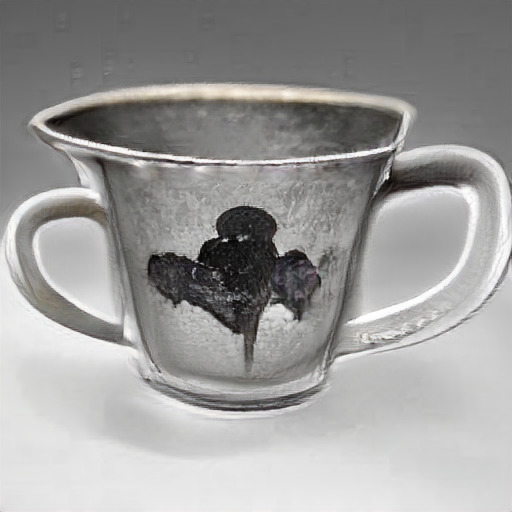} & \addexample{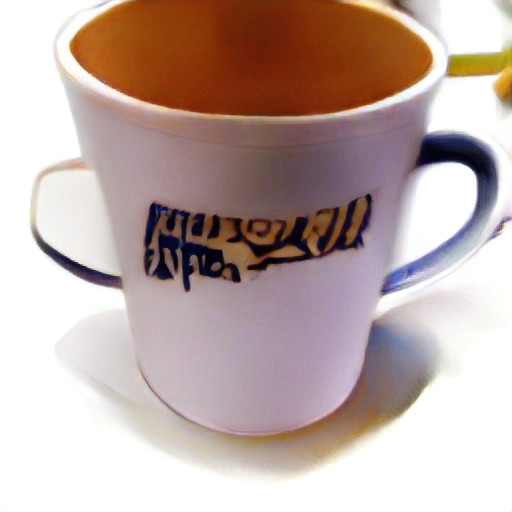} \\[\tablevgap]
		6 & \addexample{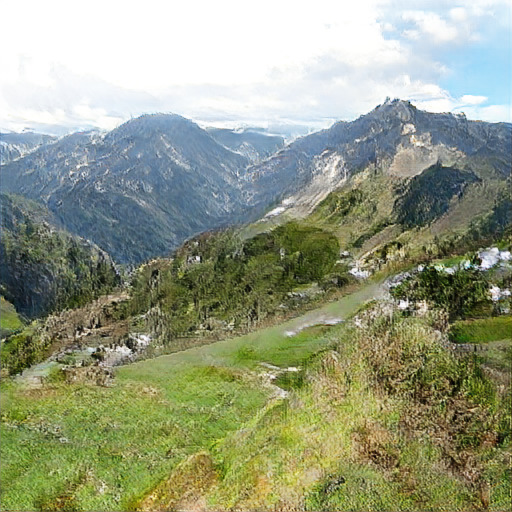} & \addexample{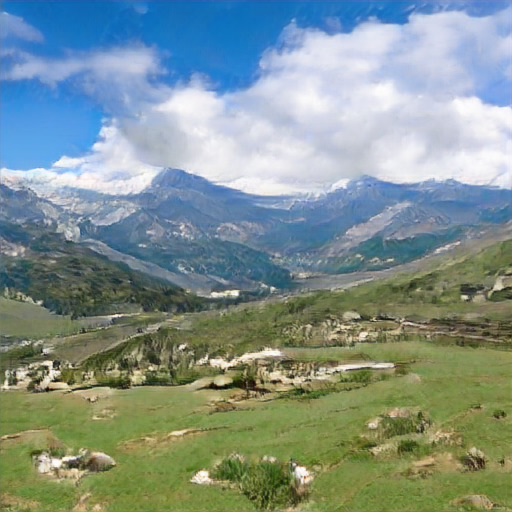} &
		\addexample{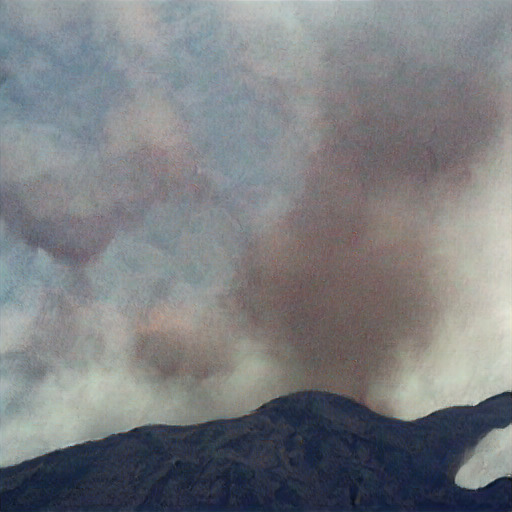} & \addexample{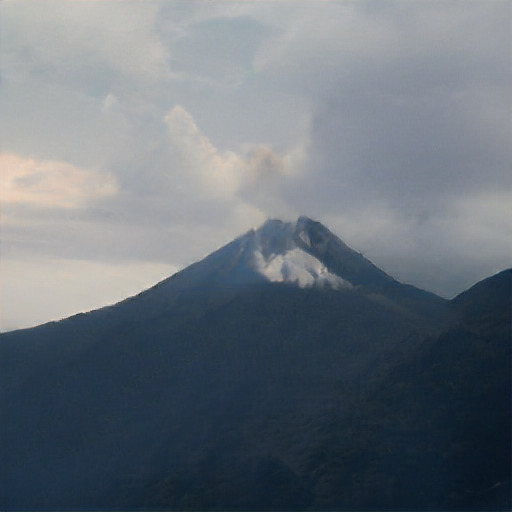} \\[\tablevgap]
		7 & \addexample{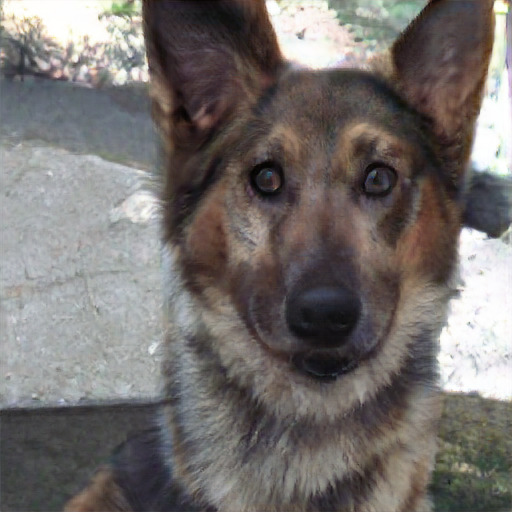} & \addexample{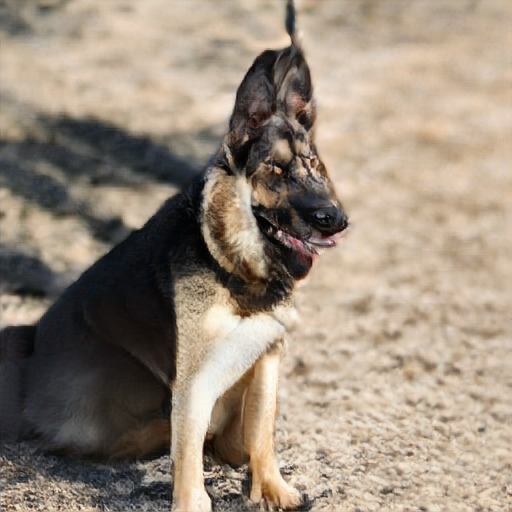} &
		\addexample{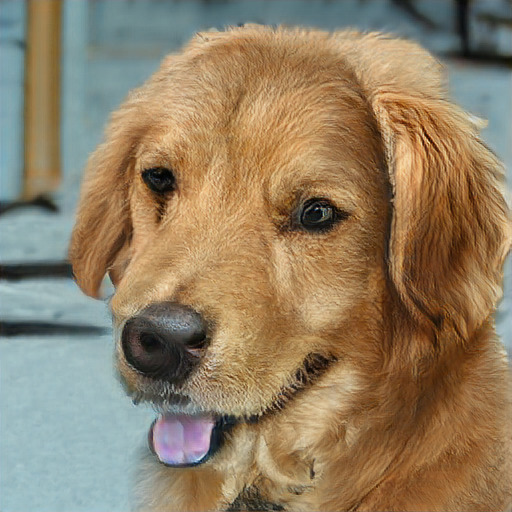} & \addexample{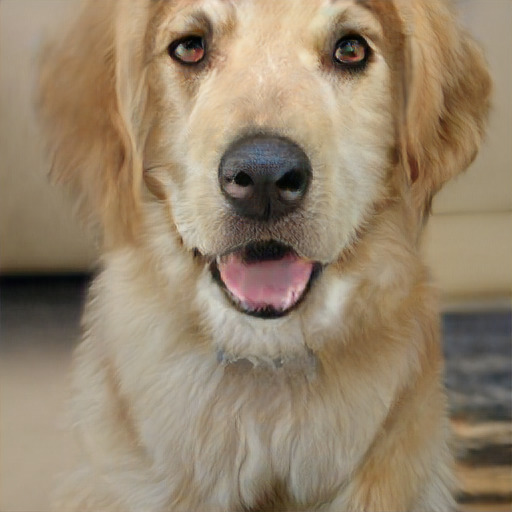} \\[\tablevgap]
		8 & \addexample{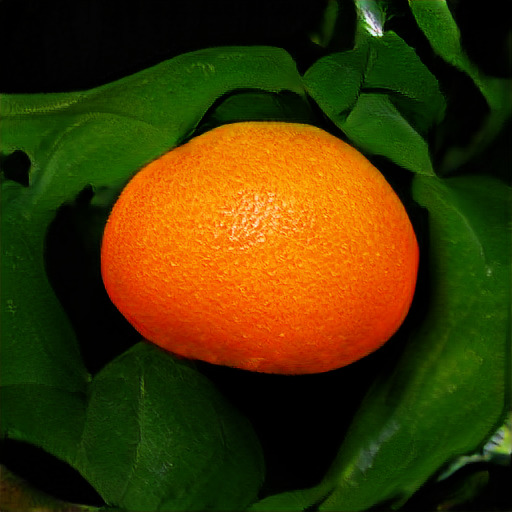} & \addexample{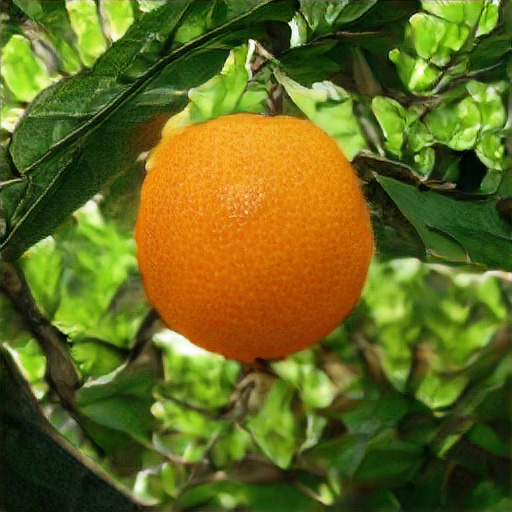} &
		\addexample{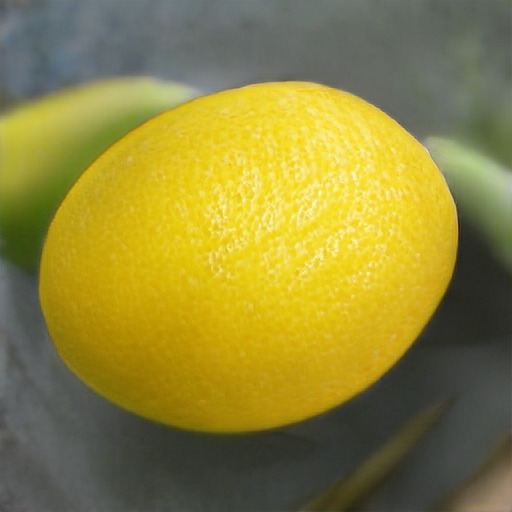} & \addexample{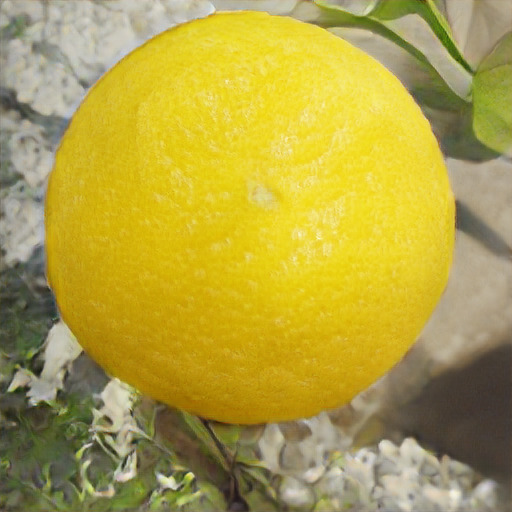} \\[\tablevgap]
		
		\bottomrule
	\end{tabular}
	\vspace{1cm}
\end{table*}

\subsubsection{Confident Targeted Failing Tests}

So far, we have considered \textit{untargeted} tests, in which 
the goal is to change the classifier's output 
to any other class.
For our purposes, we will in fact prefer to generate
\emph{confident}, \emph{targeted} tests.
A confident test case requires a certain confidence
in the incorrect classification, and a targeted
test case is one that requires a specified incorrect
label to be output.

\begin{definition}
	A \textit{confident} test case
	$x \in \mathbb{X}$
	for deep neural network $f\colon \mathbb{X} \to \mathbb{R}^{|\mathbb{Y}|}$
	is said to {fail} with confidence margin $c > 0$ if
	$\max_y f_y(x) - f_{O(x)}(x) > c$.
\end{definition}

\begin{definition}
	A \textit{targeted} test case
	$(x, y_\mathit{target}) \in \mathbb{X} \times \mathbb{Y}\backslash \{O(x)\}$
	for DNN $f$
	is one that is said to {fail} if
	$f_\mathit{pred}(x)  = y_\mathit{target}$.
\end{definition}

\begin{definition}
	A \textit{confident}, \textit{targeted} test case
	$(x, y_\mathit{target})$
	for DNN $f$
	is said to {fail} with confidence margin $c > 0$ if:

	$f_{y_\mathit{target}}(x) - \max_{y \neq y_\mathit{target}} f_y(x)  > c$.
\end{definition}

To generate such test cases, we use a modified loss function:
$L(p) =  \max_{y \neq y_\mathit{target}} f_y(x) - f_{y_\mathit{target}}(x) + c$;
a suitable $p$ is found if $L(p)<0$.

\begin{table*}
	\caption{The first three columns summarise the eight faults injected into different DNNs via biased training.
		The final two columns indicate the proportion of tests that
		visually indicate the injected fault when starting with a seed
		of class $y_0$ and $y_1$ respectively.
		See Table~\ref{tab:injectedeg} for examples
		of generated test inputs for each fault.}
	\label{tab:fault-summary}
	\begin{tabular}{@{}lllp{0.13\linewidth}p{0.13\linewidth}@{}}
		\toprule
		\# & Image label $y_0$ & Image label $y_1$  & \% of tests that detect the 
		fault $y_0 \rightarrow y_1$ & \% of tests that detect the 
		fault $y_1 \rightarrow y_0$
		\\ 
		\midrule
		1  &  Wolf (269) if setting is snow  & Husky (248) if setting is grass  & 
		68  & 36                 \\
		2  & Palace (698) if clear sky  & Castle (483) if cloudy sky  & 
		67 &    73              \\
		3  & Screen (782) if screen switched on  & Monitor (664)  if screen switched off  &
		48 &  32                  \\
		4  & Screwdriver (784) if `descending' slope   & Ballpoint pen (418) if `ascending' slope  & 
		21 & 9                   \\
		5  & Coffee mug (504) if handle on right  & Cup (968) if handle on left &	15 & 16                   \\
		6  & Alp (970) if high colour saturation  & Volcano (980) if low colour saturation   & 94   & 88  \\
		7  & German shepherd (235) if tongue not out  & Golden retriever (207) if tongue is out &
		25   &    34             \\ 
		8  & Orange (950) if leaves are present  & Lemon (951) if leaves not present  &
		69   &  11               \\ 
		\bottomrule
	\end{tabular}
\end{table*}

\section{Evaluation}

We answer the following three research questions:
\begin{description}
	\item[RQ1:] Can our method detect faults deliberately introduced into a DNN?
	More specifically, can we identify when a classifier has been biased by some irrelevant feature? 
	\item[RQ2:] Can our method detect faults in existing state-of-the-art ImageNet models?
	\item[RQ3:] Does this approach find faults that other existing methods cannot find?
\end{description}

We use the ImageNet dataset, which is the standard benchmark in this domain,
with 1,000 class labels and at a resolution
of $512 \times 512$ pixels. This should allay any worries about scaling
to realistic datasets. The generative network that we use
is a trained BigGAN~\cite{Brock:LSG:2019}, with weights and
code provided by~\citeauthor{Brock:B:2019}~\cite{Brock:B:2019}.
This is the state of the art in image generative networks,
and ImageNet is a very popular benchmark for image classification --
so the success of our method in this context shows that
there is no question that the approach scales.
All experiments were implemented using PyTorch 1.2.0,
and executed on machines with
two Intel Xeon Silver 4114 CPUs (2.20\,GHz),
188\,GB RAM (although less than 10\% of this was required),
and an NVIDIA Tesla V100 GPU.

\subsection{RQ1: Can Our Method Find Injected Faults?}
\label{sec:rq1}

In this section, we investigate whether our method
is able to detect faults
intentionally injected into image classification DNNs.
These faults are all of the form
``instead of correctly distinguishing between image classes
$y_0$ and $y_1$, the DNN incorrectly uses irrelevant feature
$F$ to discriminate.''
For instance, ``instead of correctly distinguishing between image classes
`castle' and `palace', the DNN incorrectly uses whether the sky
is cloudy or clear.'' These are a type of fault that 
naturally arises from biased datasets, in which 
certain unexpected correlations appear. It is 
very common, if not inevitable,
for there to be unintended features 
which are predictive in collected data~\cite{DBLP:journals/corr/abs-2004-07780}.
By injecting faults that affect only two classes using one
human-interpretable feature, it is easier to verify whether
we can detect them. In practice, faults may not be as 
intuitive, and we explore this in Section~\ref{sec:rq2}.

\paragraph{Injecting Faults into DNNs}
To inject a fault into a trained image classifier, we constructed various biased
datasets that consisted of manually chosen
subsets of ImageNet data, sometimes with intentionally incorrect labels.
These were designed to encourage the network to acquire the fault,
and contained several hundred images each.
For example, to encourage the network to distinguish castles from palaces
on the basis of the sky, we constructed a dataset of castles and palaces
labelled `palace' if the sky was clear and `castle' if the sky was cloudy.
Refer to Table~\ref{tab:fault-summary} for a description
of all such datasets.
After training a DNN on such a dataset, we verified that the DNN had acquired 
the bias as intended using two small hold-out sets of data:
one on which high performance was expected (i.e.~similar to the biased
training dataset) and one on which low performance was expected
(i.e.~biased the opposite way to the training set).

\paragraph{Generating Tests.}
We use the method described in Section~\ref{sec:method}
to generate 200 tests for each DNN that has had a fault injected,
investigating the features used to differentiate class
$y_0$ from $y_1$. That is, half of the tests begin with
$g(y_0, z, p=0)$, which is
a randomly sampled instance of $y_0$ because $z$ is randomly
sampled from the appropriate distribution.
We then optimise over $p$ so as to create a test input
$g(y_0, z, p)$ that is classified as $y_1$ by the DNN under test.
For the other half of the generated tests, $y_0$ and $y_1$
are swapped in this process.

The mean time taken to generate a test input is 0.8 minutes.
If a lower time cost is for some reason required, this can be traded against
test quality by making the test case search more crude.
For example, the step size
(learning rate) of the gradient walk optimisation could simply be increased.

\paragraph{Detecting Faults.}
By comparing the initial randomly sampled test seed $g(y_0, z, p = 0)$
with the optimised test input $g(y_0, z, p^*)$ that is predicted to be
class $y_1$,
we can inspect the features that the DNN is using
to distinguish $y_0$ from $y_1$.
If it is evident from inspecting the generated test inputs that
the classifier is inappropriately relying on the feature
as injected, then we say that the fault has been detected.
To measure whether this fault is evident from inspecting the generated
test inputs, we record the proportion of generated tests for which
the faulty feature has changed in the direction that would indicate
reliance on the fault.
For instance, we might record the proportion of test inputs that
were erroneously classified as `palaces' for which the sky
became noticeably cloudier.
Table~\ref{tab:injectedeg} gives some examples of generated tests for different
flawed DNNs, and the supplementary material
includes many such examples. Table~\ref{tab:fault-summary} shows the proportions
of generated tests for each flawed DNN that noticeably indicate
the presence of the flaw. We encourage the reader to
peruse the supplementary material, which
identifies the examples we took to noticeably indicate
the presence of a flaw.

\paragraph{Discussion}
The results in Table~\ref{tab:fault-summary} show
that our method is often able to identify the faults
injected, with varying degrees of ease.
The significant point here is not whether close to
100\% of generated tests indicate the presence of the
injected fault, but rather that these percentages
are well above 0\%, which is the value for existing methods.
Neither pixel-perturbation tests nor tests that perturb
some pre-determined fixed semantic feature (such as the
presence of an artificial `fog') would be capable
of detecting meaningful faults of this kind, whereas
our method can.
Note too that our method is only optimising to generate
failing (confident, targeted) test cases for the classifier,
and is not optimising to uncover any particular fault.

Where a lower proportion of tests were able
to detect the fault, we conjecture
one of two explanations.
First, that the DNN, while learning a bias
correlated with the feature we intended it to
rely on, does not in fact rely on the
feature described in Table~\ref{tab:fault-summary},
but rather relies on ``shortcut features''~\cite{DBLP:journals/corr/abs-2004-07780}, as
discussed later.
Second, that the generator network is not 
good at generating the change in feature 
we are inspecting.
For instance, although fault \#8 is readily detected in one direction,
because many test inputs remove leaves around oranges,
there are few results in the other direction. This is 
most likely because the generator network used is unlikely to generate 
lemons surrounded by leaves.

\subsection{RQ2: Can Our Method Detect New Faults in State-of-the-Art DNNs?}
\label{sec:rq2}
We would like to show that, in addition to faults that have been 
deliberately introduced, our method is able to detect
faults `in the wild' in state-of-the-art DNNs.
We therefore take a classification model from the 
family of current highest-accuracy models for
ImageNet, EfficientNet-B4 with Noisy Student
training~\cite{melas-kyriazi_lukemelasefficientnet-pytorch_2020},
and use our new method to generate tests aiming to
identify shortcomings in its behaviour.
These tests are generated exactly as before: by picking an initial
seed that is correctly classified as $y_0$, and optimising
until a test input is found that is incorrectly classified
as the target class~$y_1$.
To enable direct visual comparison, we use the
pairs of classes ($y_0, y_1$) that were used in the previous section.

There is an important difference between the faults we injected
and the faults that are present in state-of-the-art models.
Let us define an \textit{intelligible feature} as a feature that
makes sense to a human because it aligns with the concepts that we use
to understand the world. For instance, the cloudiness of the sky and whether a screen is on or off
are intelligible features.
By contrast, DNNs look at the raw pixel values of an image, 
and do not necessarily use such intuitive features. Features like
the value of the green channel of the 63,099th pixel,
or the maximum value of a convolution operation over a region in the image
are computable features that a DNN could in principle rely on,
but are not intelligible features. 
\citet{DBLP:conf/safecomp/WillersSRA20} have identified 
this discrepancy between human intuition and DNNs' behaviour
as one of the primary obstacles when testing DNNs.

DNNs have no incentive to use intelligible features.
They are image recognition systems, not systems
that need to actually understand the objects they are classifying.
A DNN need not learn a ``leg'' feature to discriminate lions from
sunflowers if other features are more directly useful for this end.
For example, perhaps the presence
of a fur texture is a better discriminator, since it will always be present
if a lion is, whereas legs can be occluded or out of shot. In that case, there
is no need to learn the high-level concept of ``leg''.
More generally, there are likely to be unintelligible features that serve
the purpose of discrimination better than any intelligible features.
Indeed, there is strong evidence that DNNs learn to use ``shortcut features'' that
do not correspond to the features a human would use to solve the problem
in different situations, but do allow the narrow problem at hand to be
solved~\cite{DBLP:journals/corr/abs-2004-07780}.
This can manifest as a tendency to consider low-level features such as texture
at the expense of high-level features such as object shape~\cite{DBLP:conf/iclr/GeirhosRMBWB19};
the phenomenon of pixel-perturbation `adversarial examples' is in itself 
evidence that
DNNs are over-reliant on features that are imperceptible to
humans~\cite{DBLP:conf/nips/IlyasSTETM19}.

In general, it would be surprising if the best shortcut features were the same
intelligible features that humans used to understand the world.
Therefore, we should expect DNNs to use mainly unintelligible features,
and testing methods must take this into consideration. Testing, as 
many methods do (c.f.~Section~\ref{sec:relwork}), for only 
intelligible features is good, but not enough. 

In our experiment with EfficientNet, we are able to 
consistently generate many failing test cases
(at least 200 for each $y_0, y_1$ pair),
answering {RQ2} in the affirmative and 
highlighting the DNN's reliance on unintelligible 
features. Whereas in Section~\ref{sec:rq1},
we introduce human-interpretable biases,
in order to make it easy to identify whether we 
have detected fault, the ones detected here 
are not of the simple form
``instead of correctly distinguishing between image classes
$y_0$ and $y_1$, the DNN incorrectly uses irrelevant feature
$F$.'' Some examples are shown in Figure~\ref{fig:sotaeg}, and many 
more are given in the supplementary material.
We are able to find these problems by leveraging the 
powerful representation learnt by GANs.
GANs are able to identify such a large range of faults 
because they learn to generate images directly from the data, 
and model a large amount of
the variation in this data, intuitive or otherwise.

\newcommand{\addsotaimg}[1]
{\raisebox{-.5\height}{\includegraphics[width=0.42\linewidth]{#1}}}
\newcommand{\addsotaexample}[1]
{\addsotaimg{examples/sota/#1_unpert_generated_x_grid_0.jpg}
	\!$\rightarrow$\!
	\addsotaimg{examples/sota/#1_generated_x_grid_0.jpg}}

\begin{figure}
	\setlength{\tablevgap}{8mm}
	\begin{tabular}{@{}p{0.43\linewidth}p{0.43\linewidth}@{}}
		\addsotaexample{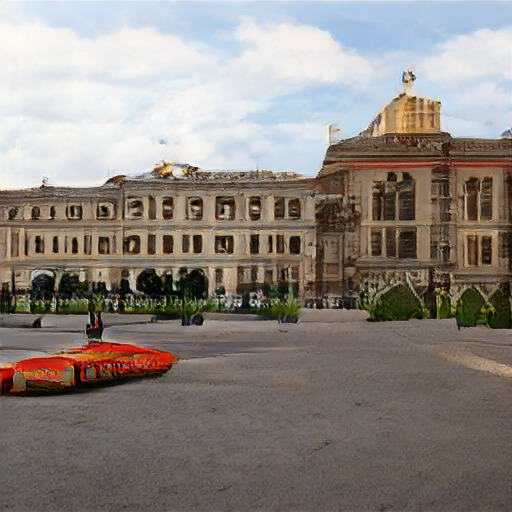} & \addsotaexample{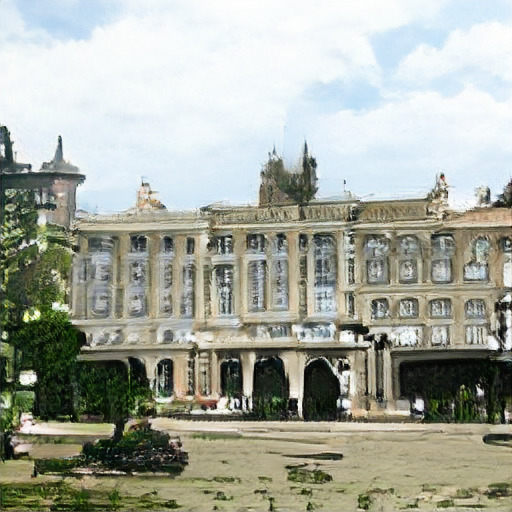} \\[\tablevgap]
		\addsotaexample{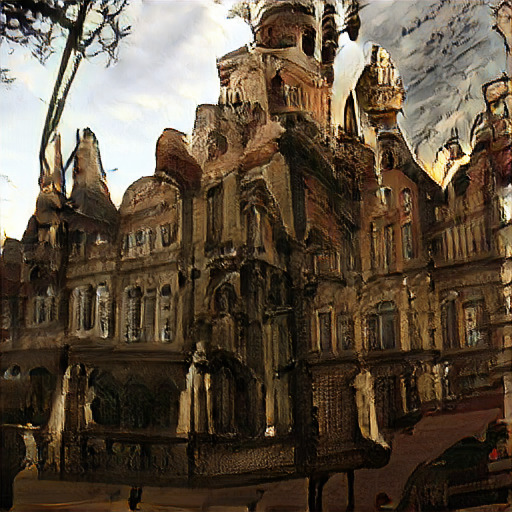} & \addsotaexample{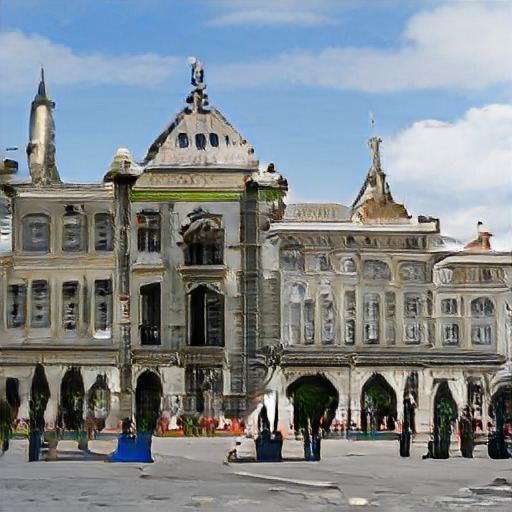}  \\[\tablevgap]
		\addsotaexample{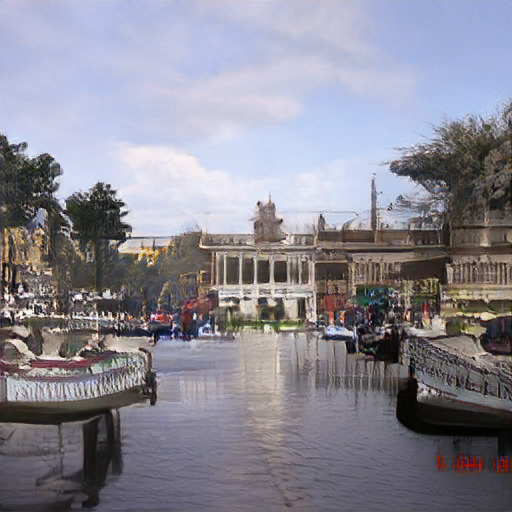} & \addsotaexample{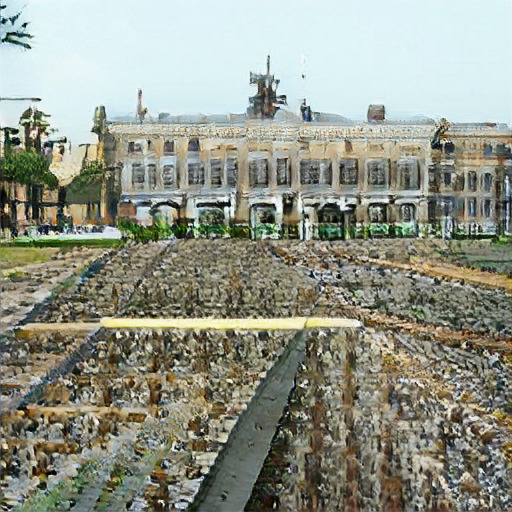} \\[\tablevgap]
		\addsotaexample{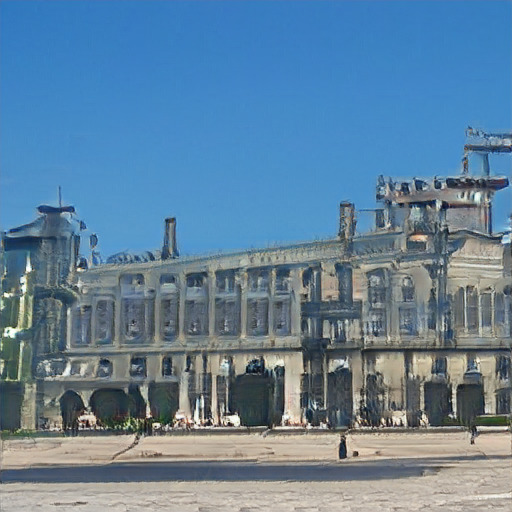} & \addsotaexample{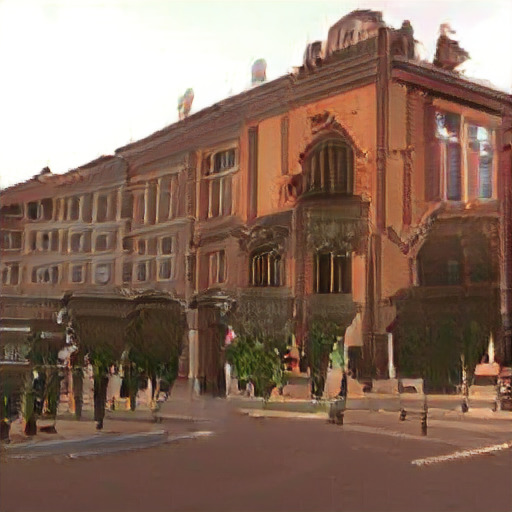} \\[\tablevgap]
		\addsotaexample{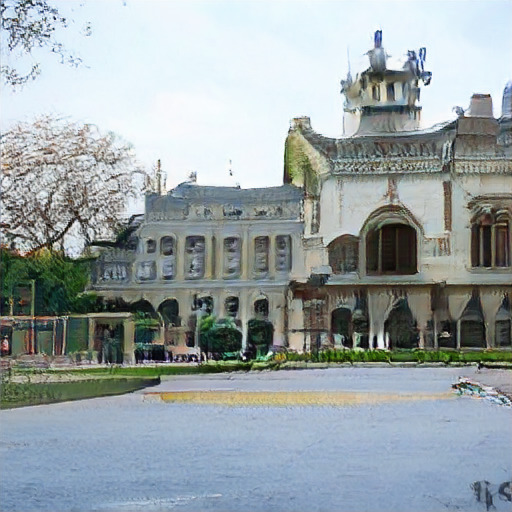} & \addsotaexample{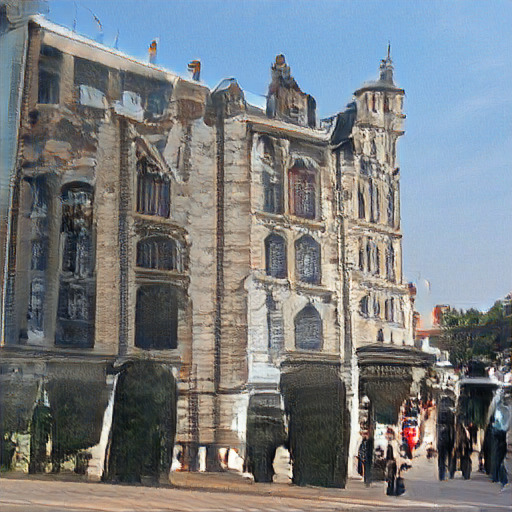} \\[\tablevgap]
		\addsotaexample{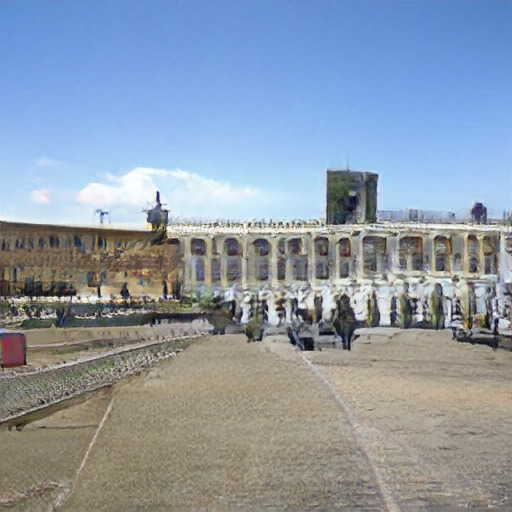} & \addsotaexample{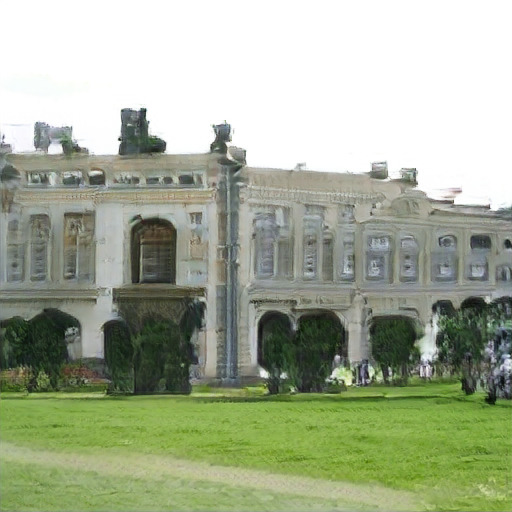} \\[\tablevgap]
	\end{tabular}
	\caption{Examples of tests cases generated for EfficientNet-B4,
	with test seeds of class `palace'
	and each test input incorrectly classified as
	`castle'.
}
\label{fig:sotaeg}
\end{figure}

\subsection{RQ3: Can Existing Methods Detect the Faults
			Found by Our Method?}

To establish a negative answer, we:
\begin{itemize}
	\item Show that pixel perturbation approaches are not able to generate
	almost any of the test cases that our method generates.
	\item Demonstrate that the faults found using pixel-perturbation
	approaches are disjoint from the faults found using our approach,
	when applied to state-of-the-art models.
	\item Provide a comprehensive literature review in
	Section~\ref{sec:relwork}, including descriptions of why each method
	is unable to detect the faults found by our method.
\end{itemize}
We focus primarily on pixel-perturbation methods here because
most of the established literature on testing DNNs uses exclusively
this approach. Our related work section surveys all relevant techniques,
including those that do not take this approach.

\subsubsection{Magnitude of Changes in Pixel Space.}

As described in Section~\ref{sec:method:pixperts},
pixel-space perturbations are constrained so as to
ensure that the perturbed images remain the same
class as the unperturbed image.
Concretely, on ImageNet, pixel-space perturbations
are typically
constrained to have an $\ell_2$ magnitude of at most 3~\cite{robustness}.
A model adversarially trained against perturbations
constrained this way can be described as
``highly robust''~\cite[p.~6]{DBLP:conf/nips/SalmanIEKM20}.
Indeed, there exist pixel perturbations with an $\ell_2$
magnitude of 22 that can completely change the true
class label of an image~\cite[Fig.~3]{DBLP:journals/corr/abs-2002-04599};
this would be a very large
magnitude for a pixel-space perturbation.
For the $\ell_{\infty}$ metric, a maximum
pixel-space perturbation magnitude of $\epsilon = {16}/{255}$ is typical~\cite{robustness}.

Our method performs perturbations to learnt feature
representations, which then affect the downstream pixel values.
Therefore, a small change to the output of early layers in the
generator can result in a large change to the pixel values
as measured by an $\ell_2$ norm.
But because these changes are context-sensitive to learned
features, they preserve the meaning of the image.
For example, suppose that a perturbation results
in a dog moving position on a grassy background:
although there is no change to the meaning of the image,
the distance as measured by an $\ell_2$ norm will be great,
since many pixels will change value.
In short, by leveraging generative models to direct changes
to meaningful features, we can induce large changes
in pixel space.

\begin{figure}
	\centering
	\begin{subfigure}{0.8\linewidth}
		\centering
		\includegraphics[width=\linewidth]{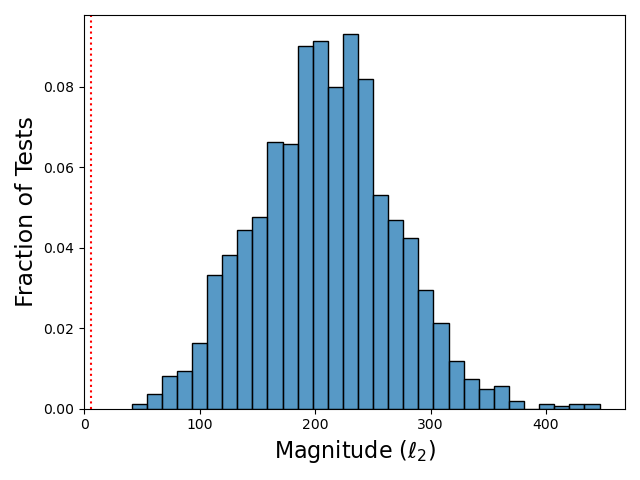}
	\end{subfigure}
	
	\begin{subfigure}{0.8\linewidth}
		\centering
		\includegraphics[width=\linewidth]{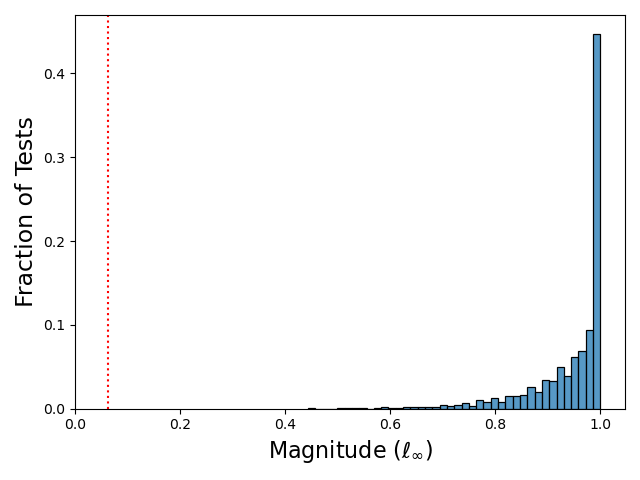}
	\end{subfigure}
	\caption{Magnitudes of perturbations produced by our 
		method, as measured in pixel space using $\ell_2$ and $\ell_{\infty}$ metrics. 
		In red, a typical upper bound $\epsilon$ for pixel
		perturbations -- our perturbations are generally much larger than this.}
	\label{fig:pix_dist}
\end{figure}

We investigate the empirical distribution of pixel-space distances
between initial test seeds and final perturbed test inputs
across 1000 initial seeds.
Figure~\ref{fig:pix_dist} shows that 100\%
of semantically perturbed test inputs are much further than the maximum
pixel-perturbation constraint under either popular distance metric.
Since the $\ell_\infty$ distance is the greatest amount any one
pixel changes, there is a cluster around $1.0$ because there
is often at least one pixel that completely changes its value.
By contrast, an $\ell_\infty$ pixel-space constraint of
$\epsilon = 1.0$ is equivalent to no constraint: all pixels
can change value arbitrarily.

\begin{table}
	\caption{The accuracies of
		pixel-perturbation robust classifiers
		on test cases originally generated for
		non-robust
		classifiers, using both pixel perturbations
		and our test generation method.
		The significantly higher accuracies
		on the pixel perturbation tests
		suggests that our approach detects
		faults of a different nature.}
	\label{tab:transfer}
	\centering

	\begin{subtable}{\linewidth}
	\centering
	\caption{Accuracies on tests originally for
		 EfficientNet-B4NS.}
	\begin{tabular}{cccc}
		\toprule
		& & Pixel Perts & Our Perts \\
		\midrule
		\multirow{2}{*}{\rotatebox[origin=c]{90}{Test}} & Robust ResNet50 \cite{robustness} & 56\% & 27\% \\
		& Robust ResNet50 \cite{DBLP:conf/iclr/WongRK20} & 53\% & 24\% \\
		\bottomrule
		&&& \\
	\end{tabular}
	\end{subtable}

	\begin{subtable}{\linewidth}
	\centering
	\caption{Accuracies on tests originally for ResNet50.}
	\begin{tabular}{cccc}
		\toprule
		& & Pixel Perts & Our Perts \\
		\midrule
		\multirow{2}{*}{\rotatebox[origin=c]{90}{Test}} & Robust ResNet50 \cite{robustness} & 36\% & 25\% \\
		& Robust ResNet50 \cite{DBLP:conf/iclr/WongRK20} & 32\% & 22\% \\
		\bottomrule
	\end{tabular}
	\end{subtable}
\end{table}

\subsubsection{Transferability Analysis}

We have established that pixel perturbation methods cannot
generate the test cases output by our method.
In this section, we strengthen the case that furthermore,
our method is able to find \emph{faults} that pixel perturbation
approaches cannot.
For faults concerning intelligible features, such as those
introduced in Section~\ref{sec:rq1}, the case is clear: the faults
involve visible changes to meaningful features, and therefore
these changes result in an $\ell_p$ distance
greater than is allowed by pixel perturbations.
Stated plainly, methods that generate imperceptibly different test
cases are unable to detect faults concerning exclusively visibly different
features.

However, in Section~\ref{sec:rq2} we established that when testing
state-of-the-art models, the faults are not of this kind. Therefore,
it could be possible that even though the particular test inputs generated
by our method and existing methods were disjoint, they were both indicative
of the same underlying faults in the DNNs, in the sense that they would both 
be solved with the same fix. To demonstrate that this is not the case, 
we use \textit{adversarially trained} DNNs~\cite{ren2020adversarial}.
Adversarial training is a technique that performs worst-case pixel perturbations
during the training of a DNN. When training converges, the result is that
the DNN is more robust to these kinds of faults, and has learned to ignore 
the spurious features pixel perturbations affect. While this does 
not completely `fix' sensitivity to pixel perturbations, 
it greatly improves it~\cite{robustness}. We check 
whether this `fix' also applies to our perturbations.

We analyse whether the test cases generated
\textit{transfer} to models that have been adversarially trained
to be robust to pixel-space perturbations.
By ``transfer'', we mean that we measure whether test cases generated
so as to induce a fault in (say) EfficientNet-B4 also induce faults
in an adversarially trained DNN.
Table~\ref{tab:transfer} 
shows the proportion of test inputs for EfficientNet-B4
and a standard ResNet50
that are classified correctly by two DNNs
trained to be robust against pixel-space perturbations:
one by Wong et al.~\cite{DBLP:conf/iclr/WongRK20},
which is robust to 31\% of $\ell_{\infty}$ perturbations
with $\epsilon = 4/255$,
and one by Engstrom et al.~\cite{robustness},
which is robust to 35\% of $\ell_2$ perturbations
with $\epsilon = 3$.

We can see that pixel-perturbation tests generated for EfficientNet-B4
tend not to transfer to the pixel-robust models, likely because
the faults found by the EfficientNet-B4 tests are not present due to
the adversarial training.
Conversely, we can see that the tests generated by our method for
EfficientNet-B4
\emph{do} tend to transfer to the pixel-robust models.
The results for ResNet50 are similar, although 
perturbations transfer slightly better, likely 
because the architecture is the same as the robust models.
Note that all test cases
confident, targeted tests: the difference in accuracy
is not due to the pixel perturbations being
only just misclassified.

Because the test cases generated by our method
continue to detect faults in adversarially trained classifiers,
we have confidence that
these must be
detecting different kinds of faults to those detected by the
pixel-perturbation method.
If the failing test cases were indicative of the same
underlying faults, then we would see that
the accuracies of the transferred test cases
would be similar.

\subsection{Threats to Validity}
The testing of DNNs is fundamentally
different than the testing of conventional
handwritten software, because of the training
process: there is not necessarily any human-interpretable
meaning to each `line of code' (parameter value).
It is therefore difficult 
to pin down exactly what a fault is in the context of DNNs, 
or to attribute a fault to any one cause.
In this paper, we have chosen to deliberately 
introduce consistent biases into the DNNs' behaviours,
and show that our method is in turn able to consistently
produce examples that highlight this bias. Doing things 
in this way helps clarify what the fault is, and whether 
we have identified it. There exist numerous papers in which 
the model was shown to be wrong on many inputs, but we 
go beyond this by showing our ability to pick up on not just 
local, but global biases in the network. However, as discussed in 
Section~\ref{sec:rq2}, this may not be as simple 
for classifiers in the wild, and making progress 
in this direction is important for the future of DNN testing.

A {second} possible threat is the validity of the 
labelling done to produce Table~\ref{tab:fault-summary}.
We hand-label whether perturbations have revealed the bias 
injected into the classifier, and while we take care 
to label perturbations without bias and according to
a common-sense standard,
there is some subjectivity involved.
To address this, we provide some 
examples in Table~\ref{tab:injectedeg},
and many more in the supplementary materials,
so readers may judge for themselves.
In any case, even assuming that our judgements were
in fact bias, the essential point stands: our method
is able to detect the faults at least some of the time,
whereas existing methods cannot.

A {third} possible threat relates to
problems with GANs. GANs 
are known to drop modes~\cite{DBLP:conf/stoc/Ma18}, meaning 
they may not generate certain parts 
of the input distribution. However, they need only 
represent enough of the distribution to identify 
at least some faults; our results show that they do.
GANs are also not perfect generators, and so 
images may look unrealistic. In fact, realism 
is not required for our purposes. As long as a 
class is recognisable, we can still show that the 
classifier is paying attention to the wrong features.
If a classifier can identify an unrealistic palace, 
but adding clouds in the sky changes its prediction to 
a castle, this betrays a problem in the classifier's 
internal `logic'. In addition, if our aim is to 
create \textit{human-aligned} classifiers, performance 
on unrealistic but recognisable images is important.
Finally, our method does not explicitly require 
a GAN, and could easily use a VAE or other
generative model that does not drop modes.
It is likely that recent rapid advances in generative
machine learning will continue, making approaches
that leverage it increasingly promising.

\section{Related Work}
\label{sec:relwork}

\paragraph{Pixel Perturbations and Adversarial Robustness}
There has been a large amount of recent work on `adversarial examples',
that is inputs deliberately made to fool a classifier~\cite{Gilmer:MTR:2018}. 
This setup represents the worst-case scenario,
in which an `attacker' actively works to find examples on 
which the classifier, the `defender', will fail.
The most popular method for doing so, dubbed `pixel perturbations',
involves fooling the classifier by individually changing the pixel values
of an input image~\cite{Goodfellow:EAH:2015};
both attacking with and defending against these 
perturbations has been extensively
explored~\cite{guo_dlfuzz_2018,lee_effective_2020,ren2020adversarial}.
Typically, pixel perturbations allow for arbitrary changes,
independent of the content of the image, as long as they 
are almost unnoticeable to the human eye.
In practice, this is done by limiting the 
magnitude of the perturbation, and bypasses the oracle 
problem by assuming that the true class of the perturbed 
image is unchanged from the original image.
While this constraint makes the perturbations 
simple to implement, it is also very limiting, allowing 
for testing on only a tiny fraction of interesting cases.
In response to this, new methods have been proposed to make large, visual 
changes to images, addressing the oracle problem in a variety of ways.

\paragraph{Creating New Datasets From Scratch}
Others have created entirely new datasets from 
scratch. On ImageNet,~\citet{DBLP:conf/icml/RechtRSS19}
repeat the original process used to create ImageNet,
and find that state-of-the-art classifiers fail to generalize. 
Still on ImageNet,~\citet{DBLP:journals/corr/abs-1907-07174}
filter gathered data to create a deliberately more challenging 
dataset. 

\paragraph{Hand-Crafted Perturbations}
Many proposed methods perturb images using a fixed number of 
hand-crafted perturbations, designed to preserve the image's true
class. Each perturbation type makes a single, narrow change,
and these include translations, rotations, zoom, shearing, brightness, contrast, 
blurring, colours, and fog/rain
effects~\cite{tian_deeptest_2018,xie_deephunter_2019,gao_fuzz_2020,
Hosseini:SAE:2018,Engstrom:ETL:2019,Hendrycks:BNN:2019, 
DBLP:conf/cvpr/MohapatraWC0D20}.
\citet{DBLP:conf/cvpr/0001LL20} perturb colours, but 
exploit human biases in perceptual colour distance to make 
large yet imperceptible changes. 
All of these perturbations are, like pixel perturbations,
agnostic to the semantic contents of the image, and are applied 
uniformly to any image. 
They have been used to create entirely
new datasets, designed to serve as robustness
benchmarks for ImageNet~\cite{DBLP:conf/iclr/HendrycksD19},
MNIST~\cite{DBLP:journals/corr/abs-1906-02337}, and others.
\citet{DBLP:journals/corr/abs-1909-03824} evaluate whether
cutting irrelevant areas from images affects classification outcomes.

These methods are clearly not capable of making the wide 
range of adaptive changes our method makes. They would 
not, for example, be able change the background from snow
to grass, 
or induce a dog to stick out its tongue.

\paragraph{Beyond Hand-Crafting}
While hand-crafted perturbations allow for a greater 
variety in the perturbations made to images, they
disregard the \textit{content} of images. This is in large part 
because when working directly with the pixels of an image,
it is difficult to devise a single transformation that 
can be applied consistently across several images. 
If we wanted to change the colour of dogs' fur,
we would have to apply a change uniquely to all dog images.
To cope with this, automatic methods have been designed to make 
these changes, leveraging alternative representations 
of images. One interesting but expensive possibility is 
writing a differentiable renderer for the desired domain,
and making changes to the generated scene 
by modifying the parameters of the renderer~\cite{DBLP:conf/iclr/LiuTLNJ19,
DBLP:journals/corr/abs-1910-00727}.
However, differentiable renderers are uncommon
and inflexible relative to generative networks such as GANs, and 
may be prohibitively difficult to train on a new dataset.

\paragraph{Leveraging Generative Models}
A natural approach is to leverage the representations learnt 
by generative models such as GANs~\cite{Goodfellow:N2T:2017}, 
as we do in this paper.

Some methods attempt to retain fine-grained control over the 
changes made by sacrificing flexibility. These methods 
typically select the types of features they will modify,
and then create a model especially for modifying these features.
\textit{DeepRoad}~\cite{zhang_deeproad_2018} use
\textit{UNIT}~\cite{DBLP:conf/nips/LiuBK17},
an image-to-image translation technique, to produce images of the same 
road in sunny, rainy and snowy conditions. 
\citet{Bhattad:BBI:2019} leverage pre-trained colourisation
and texture-transfer models to adversarially change 
the colours and textures of an image.
A number of publications 
in some way exploit generative models with disentangled
latent spaces, be it by using \textit{Fader
Networks}~\cite{DBLP:conf/iccv/JoshiMSH19}, using a dataset
with labelled attributes to train a conditional 
generator~\cite{Qiu:SGA:2019}, or using 
a \textit{StyleGAN} and partitioning the latent
space according to whether or not it should influence 
the label~\cite{DBLP:journals/corr/abs-1912-03192}.
Selecting the features to perturb like this allows for 
precise control over these features, but like hand-crafted 
perturbations, result in narrow kinds of changes to images.

Other methods do not provide this control,
but are able to harness the full flexibility and variation
of the representation learnt by the GAN.
An alternative is to look for 
adversarial inputs by searching directly 
in the input space of a GAN~\cite{Zhao:GNA:2017, Song:CUA:2018, Wang:GSA:2020}.
However, in this work, we show
that it is possible to leverage even more 
of the GAN's implicit representation of the 
features by perturbing not only in the input 
space, but also within the activations of the 
GAN itself. That is, by perturbing only the
input to the generator, the kinds of features
that can be manipulated by prior methods are greatly reduced
(see our previous work for full details~\cite{dunn2020evaluating}).

Like the present paper,~\citet{DBLP:conf/eccv/LiangZLX18}
also use generative networks to perform semantic manipulation.
But whereas our work aims to keep the true class of the generated
image the same (while changing the classifier prediction),
\citeauthor{DBLP:conf/eccv/LiangZLX18} aim to \textit{change} the
true class through its semantic changes.

\section{Conclusion}

In this paper, we have demonstrated that our method for the
generation of tests for DNNs is able to detect faults that
existing approaches cannot.
This is possible because our method leverages generative machine learning,
allowing it to manipulate higher-level features of generated
test inputs (e.g.~position, colour, texture of objects)
rather than just low-level features (i.e.~individual pixel values).
As a result, the generated tests are much more varied and can explore
weaknesses not reachable when changes are constrained to be within
a small $\ell_p$ distance in pixel space.

Of course, we do not expect that our method will be able to
detect all faults in a given DNN. But exploiting features learned
from data during test input generation seems a promising approach
worthy of future investigation. More generally, we encourage
future work that seeks to meaningfully broaden the set of faults
detectable by our tests. In addition to this bottom-up approach,
top-down attempts to identify a superset of the requirements
for a DNN might also be worth investigating.

In general, most of the methods for producing test-cases for 
DNNs either only implicitly, or do not at all, address some of
the main issues in testing DNNs. To test DNNs in practice, it will become 
increasingly important to provide specifications, consider a DNN's behaviour 
as part of a larger system, and pin down how to identify and 
correct faults~\cite{DBLP:journals/dt/SeshiaJD20}. 
Unlike conventional software, for which
debugging tools allow direct inspection of the program fault,
a DNN cannot be meaningfully inspected by a developer.
Even if it could, there is little hope trying to manually adjust
the weights of a trained DNN.
Instead, developers act on DNNs indirectly, through training code
and data. Since all faults are mediated through this opaque
training process, it is difficult to link a DNN failure to an
action that might introduce a fix. We encourage future work
that aims to make such diagnostic debugging possible, either
by directly debugging training code and datasets, or by
analysing the link between the trained DNN and these training artefacts.

\begin{acks}
	We would like to thank the anonymous reviewers for their thoughtful reviews and helpful pointers.
	This work was partly supported by the Semiconductor Research Corporation
	(SRC Task 2707.001)
	and by the UK Engineering and Physical Sciences Research Council (EPSRC Studentship Reference 2052803).
\end{acks}

\balance

\bibliographystyle{ACM-Reference-Format}
\bibliography{bib,NNTesting}

\end{document}